\documentclass[letterpaper, 10 pt, conference]{ieeeconf}
\IEEEoverridecommandlockouts%
\usepackage[utf8]{inputenc}
\usepackage{booktabs}
\usepackage{amsmath, amssymb}
\usepackage{graphicx}
\usepackage{listings}
\usepackage{algorithm, algorithmicx}
\usepackage[noend]{algpseudocode}
\usepackage{fancyvrb}
\usepackage{subcaption}

\usepackage{hyperref}

\begin{document}

\title{\LARGE \bf ProbRobScene: A Probabilistic Specification Language \\ for 3D Robotic Manipulation Environments}
\author{Craig Innes and Subramanian Ramamoorthy*%
\thanks{*This work is supported by funding from the Alan Turing Institute, as part of the Safe AI for Surgical Assistance project.}
}

\maketitle
\thispagestyle{empty}
\pagestyle{empty}

\begin{abstract}
    Robotic control tasks are often first run in simulation for the purposes of verification, debugging and data augmentation. Many methods exist to specify what task a robot must complete, but few exist to specify what range of environments a user expects such tasks to be achieved in. ProbRobScene is a probabilistic specification language for describing robotic manipulation environments. Using the language, a user need only specify the relational constraints that must hold between objects in a scene. ProbRobScene then automatically generates scenes which conform to this specification. By combining aspects of probabilistic programming languages and convex geometry, we provide a method for sampling this space of possible environments efficiently. We demonstrate the usefulness of our language by using it to debug a robotic controller in a tabletop robot manipulation environment.
\end{abstract}

\section{Introduction}
\label{sec:intro}

Consider a typical robotic manipulation scenario one might encounter in a factory - picking up objects from an in-tray and placing them in an out-tray \cite{zeng2018robotic}. Designers often first test such controllers in a simulator (e.g., PyBullet \cite{coumans2019} or CoppeliaSim \cite{rohmer2013coppeliasim}) to efficiently verify that the controller accomplishes their chosen pick-and-place task across many variations in the environment. However, while many methods exist to declaratively specify what \emph{task} a robot must accomplish, few options exist for specifying the \emph{range of environments} the robot is expected to accomplish that task in. This is despite the fact that adequately defining the intended operational environment of a robot is a crucial component of system validation \cite{fratrik2019many}.

A typical way to test a robot works in multiple environments is to explicitly construct each simulation scene --- The user manually places the robot, table, trays, and other objects, runs the simulation using their robot controller, then makes ad-hoc changes to the scene to see if the controller can handle variations. This approach mirrors \emph{exploratory testing} in traditional programming \cite{itkonen2005exploratory}. Such manual tweaking is time-consuming and lacks rigour. A better approach would be to have the equivalent of \emph{automated randomized testing} \cite{godefroid2008grammar} for robotic environments. For example, in a tool like Haskell Quick-Check \cite{claessen2011quickcheck}, a user provides constraints on the input of a function and a desired test condition. Quick-check then generates 100 samples from the constrained input space and reports how many satisfy the test condition.

In our Quick-Check analogy, the input constraints correspond to constraints on valid environment setups in which our robot is expected to operate. For robotic manipulation tasks, these environmental constraints typically take the form of \emph{geometric relations} between objects. Following the goals of End-User-Programming \cite{ko2011state}, we would like users to be able to easily express their desired environmental constraints without deep knowledge of the internals of a system. A small number of such specification languages exist already, for domains such as 2D autonomous vehicle simulation \cite{fremont2018scenic}. Our aim is to build a similar language for the three-dimensional, inherently relational domain of robotic manipulation.

\begin{figure}
    \centering
    \includegraphics[width=0.7\linewidth]{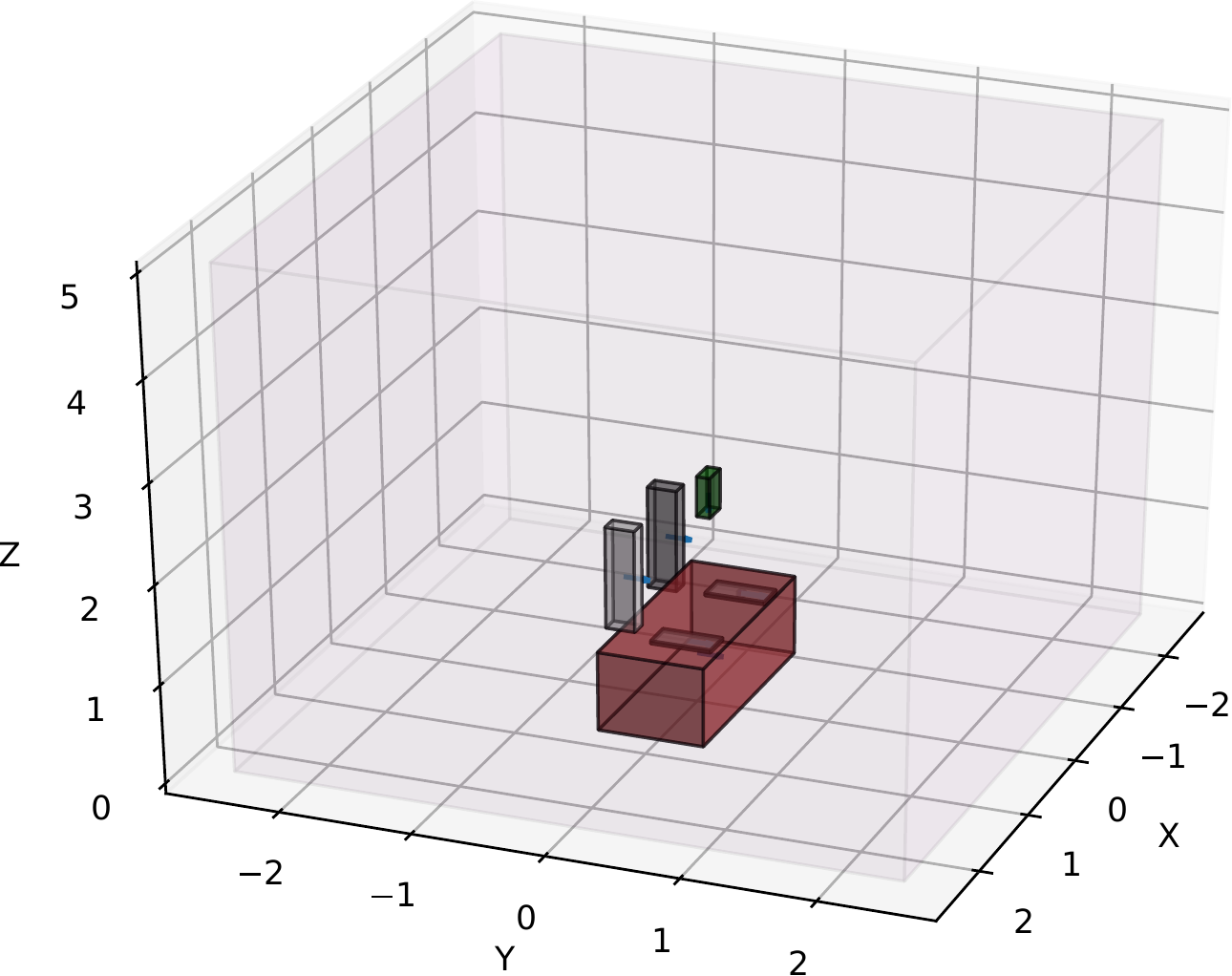}
    
    \begin{LVerbatim}[fontsize=\footnotesize]
    
t = Table on V3D(0,0,0)
r1 = Robot on (top back t) - V3D(0.4, 0, 0)
r2 = Robot on (top back t) + V3D(0.4, 0, 0)

tr_1 = Tray completely on t, 
            ahead of r1,
            left of t
            
tr_2 = Tray completely on t,
            ahead of r2,
            right of t

Cube completely on tr_1

Camera at V3D((-0.1, 0.1),
              (-0.1, 0.1), 
              (1.9, 2.1)),
       facing V3D(0, 0, -1)
    \end{LVerbatim}
    \caption{``Table-Cube'' specification -- two robots at either side of a table, two trays with a cube in one, and a ceiling camera facing downwards.}
    \label{fig:lang-example}
\end{figure}

We present ProbRobScene --- a probabilistic specification language for describing 3D robotic manipulation environments. Figure (\ref{fig:lang-example}) shows an example specification for a table-top environment, along with a sample from this specification. By combining techniques from probabilistic programming and convex computational geometry, ProbRobScene can efficiently generate sample environments consisting of complex relations between objects in three dimensions.

In subsequent sections we first briefly summarize the syntax and semantics of ProbRobScene. In particular, how the language implicitly translates declarative geometric relations into concrete regions of sample space (Section \ref{sec:prob-language:regions}). We then describe how we leverage results from convex geometry to prune the space of feasible environments, and make an efficient, domain-specific sampler (Section \ref{sec:sampling}). We conclude with a case study to show how ProbRobScene can be used to discover bugs in a two-robot handover task.

For this paper, we focus on the use-case of debugging a fixed robotic controller, but there are multiple other applications of our probabilistic environment specification language. In robot learning, ProbRobScene may be used as a data-augmentation procedure to gather targeted training data \cite{tobin2017domain}, or combined with adversarial learning techniques to teach a system to be robust to difficult edge-cases \cite{dreossi2018counterexample}.

\section{The ProbRobScene Language}
\label{sec:prob-language}

We first give a brief overview of the main features of a ProbRobScene specification, then explain how to translate these specifications into a form we can sample.

To grasp the basics of the language, consider the line:

\begin{LVerbatim}
c1 = Cube with width 0.8,
          with length (0.2, 0.4)
\end{LVerbatim}

Here, we create an object of type \texttt{Cube} (upper-case) and assign it to variable \texttt{c1} (lower-case). Object types such as \texttt{Cube} are defined in a separate \texttt{models} file, whose definitions are then imported into the current specification. Objects are defined in a syntax similar to Python classes:

\begin{LVerbatim}
class Cube:
  width: 0.1
  length: 0.1
  height: 0.1
  color: 'gray'
\end{LVerbatim}

In addition to the custom properties set in their definition, all object classes inherit a \texttt{position} and \texttt{orientation} property. Following creation, we then append a comma-separated list of statements to the object which modify its default properties. The arguments to these statements can either be concrete values (e.g., set \texttt{width} to \texttt{0.8}) or a \emph{distribution} over a range of values (e.g., set \texttt{length} to the uniform distribution between \texttt{0.2} and \texttt{0.4}). ProbRobScene must sample such distributions to produce a concrete scene.

These modifying statements are called \emph{specifiers}, and build on the syntax described in the Scenic language \cite{fremont2018scenic}. The specifier syntax allows the user to arbitrarily order their modifications to an object, with the language internally resolving sampling dependencies between properties using a topological sorting algorithm. For example, the line:

\begin{LVerbatim}
c2 = Cube facing towards c1,
  with position V3D((0.5, 1.0), 0, 0)
\end{LVerbatim}

Rotates \texttt{c2} to face towards the position of \texttt{c1}. However, this rotation \emph{depends on the position of c2}, so internally, the language first samples c2's position before calculating its orientation. For a detailed explanation of how specifier dependency resolution is implemented, see \cite{fremont2018scenic}.

\subsection{Semantics of Simple Specifiers}

Table (\ref{tab:specifiers}) summarizes the specifiers available in ProbRobScene. The left column shows the syntax of each specifier, the object property it applies to (e.g., position, orientation, or any), and the data type of the arguments it takes. The right column shows the type of value the given specifier evaluates to. For most simple specifiers, this value is a three-dimensional vector (V3D). We summarize the semantics of each of these simple specifiers below:

\paragraph*{\textbf{with}} General purpose direct assignment. For example, \texttt{Cube with color "green"} sets the cube's \texttt{color} property to \texttt{"green"}.
\paragraph*{\textbf{facing / facing towards}} Rotates the object to face either a given normal vector, or towards another object. For example \texttt{Cube facing V3D(1, 0, 0)}, rotates the cube 90 degrees to face forwards along the X-axis.\footnote{This assumes the canonical ``forward'' direction is the positive y-axis.}
\paragraph*{\textbf{at}} Sets the position of an object to a given vector.
\paragraph*{\textbf{beyond}} Best explained by example --- The statement \texttt{Cube beyond X by d from Y} sets the cube position \texttt{d} metres from \texttt{X} along the direction vector from \texttt{Y} to \texttt{X}.

\newcommand{\ra}[1]{\renewcommand{\arraystretch}{#1}}
\begin{table*}
    \caption{Specifier syntax \& return assignment type}
    \ra{1.1}
    \centering
    \begin{tabular}{@{}ll@{}}
    \toprule
         Specifier Syntax & Return Assignment Type \\
    \midrule
         \textbf{Position Specifiers}& \\
         at \emph{V3D} $[\text{relative to } (\textit{V3D} \mid \textit{Object})]$ & \emph{V3D} \\
         beyond (\emph{V3D} $\mid$ \emph{Object}) by $(\textit{Float} \mid \textit{V3D})$ from $(\textit{V3D} \mid \textit{Object})$ & \emph{V3D} \\
         in \emph{Region} & \emph{Region-Dist}\\
         $(\text{left} \mid \text{right} \mid \text{ahead} \mid \dots)$  of $(\emph{Object} \mid \textit{V3D})$ & \emph{Halfspace-Dist}\\
         $[ \text{completely} ]$ on $(\textit{Object} \mid \textit{Rectangle-Region} \mid \textit{V3D})$ & $\textit{V3D} \mid \textit{Rectangle-Dist}$ \\
         aligned with $(\textit{Object} \mid \textit{V3D})$ on $(x \mid y \mid z)$ & \emph{Rectangle-Region} \\
    \midrule
         \textbf{Orientation Specifiers}& \\
         facing \emph{V3D} & \emph{V3D}\\
         facing Toward $(\textit{Object} \mid \textit{V3D})$ & \emph{V3D}\\
    \midrule
        \textbf{Generic Specifiers} & \\
         with \emph{String} \emph{Any} & \emph{Any} \\
    \bottomrule
    \end{tabular}
    \label{tab:specifiers}
\end{table*}

\subsection{Relational Specifiers Defining Regions}
\label{sec:prob-language:regions}

Rather than specifying a property to be a single vector, the remaining specifiers assign a property a \emph{distribution} of points over an entire geometric region. Users can define custom regions explicitly using one of several \texttt{Region} constructors, given in Figure (\ref{fig:regions}).

\begin{figure}
\begin{LVerbatim}[fontsize=\footnotesize]
Cuboid(origin, orientation, dims)
Rect3D(origin, orientation, dims)
Halfspace(origin, normal)
ConvexPolygon3D(hsi, origin, normal)
ConvexPolyhedron(hsi)
All()
Empty()
\end{LVerbatim}
\caption{Region Constructors}
\label{fig:regions}
\end{figure}

The most direct way to set an object's position to a region distribution is with the \textbf{in} specifier. This specifier simply produces a uniform distribution over all points contained within the given geometric region.

However, rather than defining their own regions explicitly, an easier way for the user to specify their environment is by producing regions \emph{implicitly} via other relations. 

First, consider \texttt{Cube ahead of X}. This allows the cube to be placed anywhere in the world, as long as this position is in front of X. This is geometrically equivalent to saying the position the must lie within the \emph{half-space} with an origin and normal given by the front face of X.

Next is \texttt{Cube on X}. It says that the cube is placed somewhere in the space on top of X. Geometrically, this corresponds to a rectangular region defined by the top face of X. Notice though that there are points on the surface of X for which the edges of the cube might hang off the edge of X. So optionally, the user can use the \texttt{completely} keyword to assert that the full bounding box of the cube must be inside the surface rectangle of X.

Finally, lets look at \texttt{Cube aligned with X along z}. Geometrically, this restricts the z component of our cube to equal the z component of \texttt{X}. Thus the available positions of our cube are given by a plane with origin $\langle 0,0,X.z \rangle$ and normal $\langle 0, 0, 1 \rangle$. 

Note that (unlike Scenic \cite{fremont2018scenic}) we are not restricted to a \emph{single} positional specifier per object. In many situations, the most direct way to express the possible positions of an object is as a \emph{combination} of multiple relations. For example, a Cube could be \textbf{on} the floor \textbf{in front of} the table. We will see in the next section how this complicates the problem of sampling the space, and how we resolve it.

\section{Efficient Generation of Sample Scenes from a ProbRobScene Specification}
\label{sec:sampling}

Given a ProbRobScene specification, we now want to sample scenes which conform to it. If every property in our specification were given by an unconstrained uniform distribution, this would be trivial---sample each object property in dependency order. However, given the specifiers discussed above, we know that many properties may have distributions composed of multiple overlapping geometric regions.

As a naive approach, we could try rejection sampling. Following this method, we would treat each relational specifier as a constraint on the given property. We first sample a value of each property from the entire input space. If this sample conforms to all constraints, we accept it. Otherwise, we reject this sample and try again.

Naive rejection sampling is often effective in two dimensions. For problems in three dimensions (or higher) though, we face the curse of dimensionality: When more than a few geometric constraints are specified, the volume of valid distribution space becomes a tiny fraction of the total space. This results in an unacceptably large number of rejections before a valid sample is found. So we must take this problem seriously by first pre-calculating the regions resulting from combinations of relational constraints. Then, we must come up with an efficient method for sampling regions corresponding to arbitrary convex polyhedrons.

\subsection{Multiple specifiers as Half-space Intersection}

Imagine we have a cube with three position specifiers---It must be \textbf{in} some rotated cuboid, it must be \textbf{in front of} the table, and it must be \textbf{on} the floor (an axis-aligned box). An acceptable point must land within the regions of all three specifiers. Geometrically, the position of the cube must lie within the \emph{intersection} of these three regions.

Ideally we would like to have a way of transparently computing any combination of regions from Figure (\ref{fig:regions}) with a unified representation, without using specialised methods to accommodate the specific details of each region type. Fortunately, since all of our region types represent convex shapes, we can use a result from convex algebraic geometry: \emph{Any} convex shape can be represented as an \emph{intersection of half-spaces} \cite{boyd2004convex}. Formally, an intersection of half-spaces is a set of points $x$ which obey the constraints:

\begin{equation}
\label{eqn:hsi}
    Ax \leq b
\end{equation}

Here, $A$ is a $m \times d$ matrix, where $m$ is the number of intersecting half-spaces, and $d$ is the dimension of $x$. $b$ is $m$-dimensional vector. As an example, consider an axis-aligned cuboid region with size $3 \times 2 \times 1$,  centred at $\langle 0.5, 0.5, 0.5 \rangle$ its half-space representation is:

\begin{equation}
    \label{eqn:hsi-example}
    \begin{bmatrix}
    -1 & 0 & 0 \\
    1 & 0 & 0 \\
    0 & -1 & 0 \\
    0 & 1 & 0 \\
    0 & 0 & -1 \\
    0 & 0 & 1
    \end{bmatrix}
    x +
    \begin{bmatrix}
    3.5 \\ -2.5 \\ 2.5 \\ -1.5 \\ 1.5 \\ -0.5
    \end{bmatrix}
    \leq \vec{0}
\end{equation}

This representation has two immediate advantages: First, finding the intersection of two shapes becomes trivial. If we have a region $R_x = \{x \mid A x \leq b\}$, and another region $R_y = \{y \mid C y \leq d \}$, then their intersection $R_x \cap R_y$ is just the union of constraints comprising each halfspace:

\begin{equation}
\label{eqn:hsi-intersection}
R_x \cap R_y = \{z \mid \begin{bmatrix} A \\ C \end{bmatrix} z \leq \begin{bmatrix} b \\ d \end{bmatrix} \}
\end{equation}

Second, checking whether a region contains a point $x$ is also trivial---just check whether $x$ satisfies equation (\ref{eqn:hsi}).

\subsection{Object containment as morphological erosion}

An additional complication is how we compute relations which assert that an object must be completely contained within a region. Recall this is different from saying that only the position must lie within a specified region---\emph{all} points on the object must lie within the region. For a region defined by points in $R_x$ and an object by $R_y$, containment relations are defined by a morphological erosion:

\begin{gather}
\label{eqn:erosion}
R_x \ominus R_y = \{z \mid \forall y \in R_y, z + y \in R_x\}
\end{gather}

Fortunately, with our representation of objects and regions as half-space intersections, equation (\ref{eqn:erosion}) is \emph{also} a half-space intersection, given by equation (\ref{eqn:erosion-hsi}):

\begin{gather}
\label{eqn:erosion-hsi}
    R_x \ominus R_y = \{z | Az \leq b - e^* \} \\
    e^*_i = \max_{y \in R_y} (A_i \cdot y)
\end{gather}

The value of (\ref{eqn:erosion-hsi}) is computable by a simple linear program.

\subsection{Sampling Arbitrary Convex Regions}

We've now managed to compactly represent all our combinations of relational specifiers as intersections of half-spaces. Continuing our philosophy of injecting domain-specific knowledge into the probabilistic programming paradigm, we now leverage this geometric structure to efficiently sample from our specification.

At this point, we could again attempt to naively sample from the bounding box of our convex regions, then reject those samples which fall outside of the half-space intersection. Again, in three dimensions, the volume of a half-space intersection is likely to be a tiny fraction of the volume of its bounding box, resulting in a unacceptably high rejection rate. Instead, we re-purpose an algorithm for sampling convex meshes called hit-and-run sampling \cite{chalkis2020volesti}.

Algorithm (\ref{alg:hit-and-run}) outlines the main steps: Starting from a feasible point $p_0$ within the half-space intersection $R$, draw a line in a random direction $\vec{d}$ through $p_0$ that intersects the boundaries of the half-space intersection at points $p_0 + m_a * \vec{d}$ and $p_0 + m_b * \vec{d}$. Then, pick a point $p_1$ random on that line segment. Repeat the previous steps starting from the new point $p_1$. This procedure is guaranteed to yield a uniform point from the region $R$ with a mixing time of $O(d^3)$, where $d$ is the dimensionality of the input space.

\begin{algorithm}
        \caption{Sampling of Half-space Intersection with Hit And Run}
        \label{alg:hit-and-run}
        \label{alg:full-system}
        \begin{algorithmic}[1]
        \Function{HitAndRun}{HSI, $p_0$, max-iterations}
            \For{$i = 1$ to $\text{max-iterations}$}
            \State $d \sim  \langle \mathcal{N}(0, 1), \mathcal{N}(0, 1), \mathcal{N}(0, 1) \rangle$
            \State $d \gets d / |d| $
            \State Get $m_a \geq 0$ where $p_{i - 1} - m_a * d$ on HSI bounds
            \State Get $m_b \geq 0$ where $p_{i - 1} + m_b * d$ on HSI bounds
            \State m $\sim$ \Call{Uniform}{$-m_a$, $m_b$}
            \State $p_{i} \gets p_{i - 1} + md$
            \EndFor
        \State \Return x
        \EndFunction
        \end{algorithmic}
\end{algorithm}

We now have a comprehensive method for taking a full ProbRobScene specification and efficiently sampling from it. This method has some limitations. A key limitation is that we still must reject a sample if there is a collision between objects in overlapping regions. This is because we do not pre-compute the parts of a region in which there would be space to place the remaining objects. The subset of a region which is not occupied by other objects is potentially non-convex, so does not fit neatly within our formulation. These issues are outside the scope of this paper, but calculating such regions is related to the \emph{3D Nesting Problem} \cite{rocha2019robust} in constraint programming, and may make for interesting future work.

\section{Case Study: ProbRobScene for Debugging}
\label{sec:case-study}

We now show an example of how ProbRobScene can be used to quickly and automatically check an existing robot controller for bugs by taking our ``Table-Cube'' specification from Figure (\ref{fig:lang-example}) through to the simulation stage.

\begin{figure*}
    \centering
    \begin{subfigure}[t]{0.32\linewidth}
        \includegraphics[width=\textwidth]{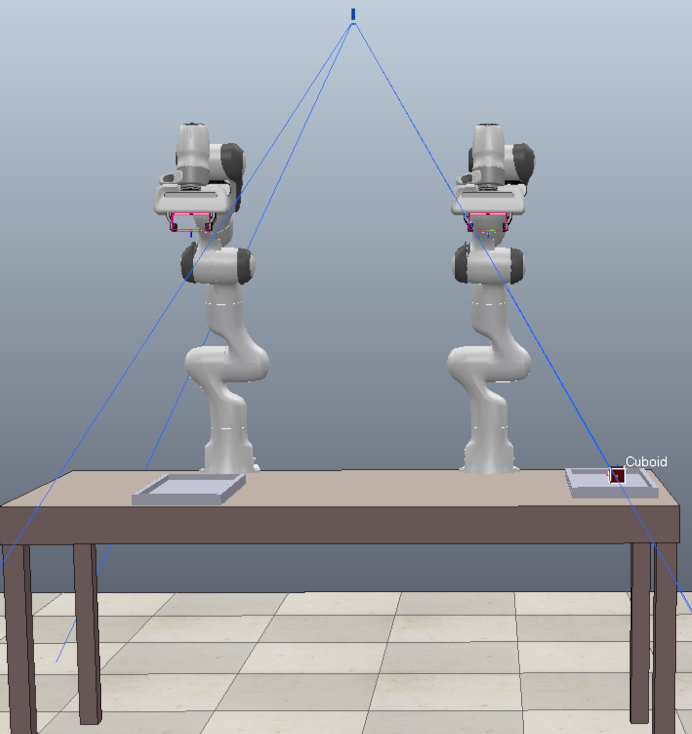}
        \caption{Perception Fault}
        \label{fig:coppelia-ex:perception}
    \end{subfigure}
    \begin{subfigure}[t]{0.32\linewidth}
        \includegraphics[width=\textwidth]{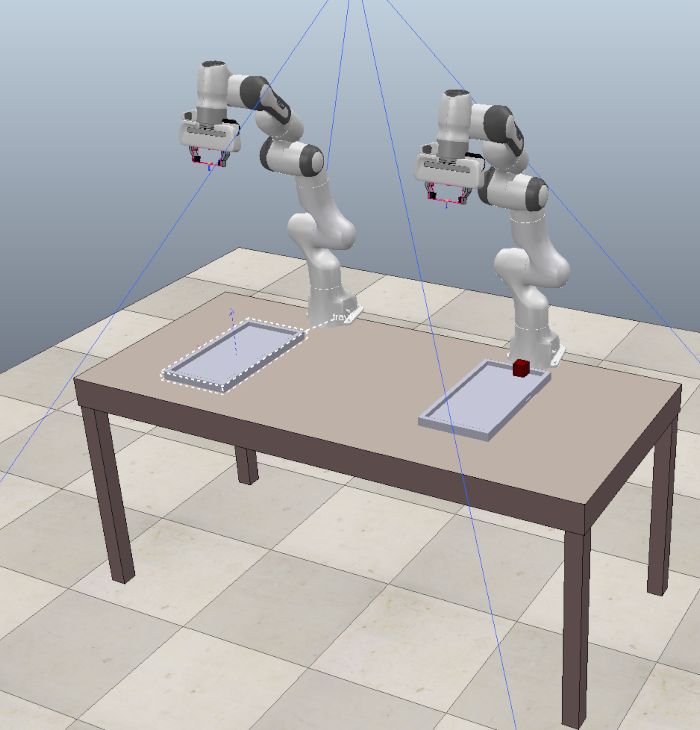}
        \caption{No Path}
        \label{fig:coppelia-ex:no-path}
    \end{subfigure}
    \begin{subfigure}[t]{0.32\linewidth}
        \includegraphics[width=\textwidth]{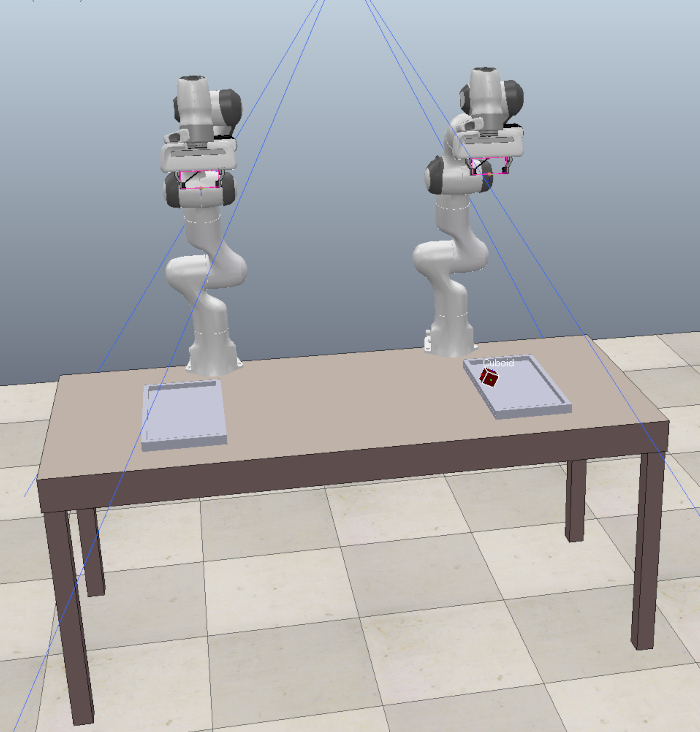}
        \caption{Poor Grip}
        \label{fig:coppelia-ex:bad-grip}
    \end{subfigure}
    
    \caption{Samples from ``Table-Cube'' specification simulated using Coppelia-Sim. One of each failure type is shown.}
    \label{fig:coppelia-ex}
\end{figure*}

The ``Table-Cube'' task is as follows: we have  two Franka Emika Panda robot arms on either side of a table, a ceiling mounted RGB-D camera, two trays (one within reach of each robot), and a cube on the left tray. The goal is to move the cube from the left tray to the right tray. Figure (\ref{fig:coppelia-ex}) shows samples from the ``Table-Cube'' specification loaded in the CoppeliaSim \cite{rohmer2013coppeliasim} robotic simulator. To load a ProbRobScene sample into our simulator, we need only a small wrapper class which parses the given scene information and calls the relevant object creation APIs within the simulator.

Our robot controller uses a standard open-loop planning pipeline. First, the robot takes the RGB-D image from the ceiling camera, and uses an off-the-shelf blob-detection algorithm to capture the location of the cube in world-space. Next, using a built-in inverse kinematics solver, it plans a trajectory to move the left-arm gripper just above the cube, grasps the cube, then drops it within reach of the right arm. The right arm then picks up the cube in a similar manner, and drops it into the right tray. The full code for language, sampler, wrapper, and controller is available at: \url{https://github.com/craigiedon/ProbRobScene}.

We simulate the above controller in 100 environments sampled from the ``Table-Cube'' specification, and record whether the robot successfully accomplished the task. We mark a simulation as a success if the cube ends up in the right-hand tray by the end of the simulation. If the simulation is a failure, we annotate the reason for failure. To show the benefits of the sampling techniques described in section \ref{sec:sampling}, we also sample 100 scenes using standard rejection sampling (where each relational specifier in the original specification is rephrased as a rejection constraint). Table (\ref{tab:average-rejects}) shows the average number of rejected samples per valid sample. Even with only a few constraints on a subset of scene objects, the average number of rejections exceeds 200. We did not include the number of rejections for the ProbRobScene sampler, as no samples were rejected across the experiments.

\begin{table}
    \caption{Naive Rejection Sampling on ``Table-Cube''}
    \ra{1.1}
    \centering
    \begin{tabular}{@{}ll@{}}
    \toprule
    Reject Type & Rejections (Avg of 100)\\
    \midrule
    Object Collision & 98.45 \\
    Relative Position Violation &  107.95 \\
    Containment Violation &  6.45 \\
    \textbf{Total} & \textbf{212.85} \\ 
    \end{tabular}
    \label{tab:average-rejects}
\end{table}

Table (\ref{tab:sim-results}) shows the successful and unsuccessful attempts across 100 simulations. While the robot was successful in most scenes (74), our system sampled a number of situations from our ProbRobScene specification in which the robot failed to accomplish the task. In 13 scenes (labelled ``Perception Fault'', Figure (\ref{fig:coppelia-ex:perception})) the sampler produced scenes in which the cube was obscured from camera view by a part of the arm, or where the cube was so far to the left of the table that it was partially out of the camera's viewing range. As a result, the robot's object detector failed to identify the correct location of the cube and did not manage to grasp it. In 8 scenes (labelled ``No Path'', Figure (\ref{fig:coppelia-ex:no-path})), the sampler produced scenes where the cube was either so far away from the gripper or so close to the arm's base that the inverse kinematics solver could not find a valid path to the top of the cube. In 5 cases (labelled ``Poor Grip'', Figure(\ref{fig:coppelia-ex:bad-grip})), the sampler produced scenes where the cube was placed up against a corner of the tray such that the robot's grip attempt was compromised.

\begin{table}
    \caption{Results of Simulating 100 ``Table Cube'' scenes}
    \ra{1.1}
    \centering
    \begin{tabular}{@{}ll@{}}
    \toprule
    Task Result & Quantity\\
    \midrule
    Success & 74 \\
    Perception Fault &  13\\
    No Path & 8 \\
    Poor grip & 5\\
    \end{tabular}
    \label{tab:sim-results}
\end{table}

We can see from these results that, even for a relatively straightforward controller in an environment with restricted amounts of variation, ProbRobScene can be useful for identifying potentially non-obvious edge-cases in which that controller may fail.

\section{Related Literature}

There have been numerous probabilistic programming languages in recent years which allow users to declaratively specify tasks such as probabilistic inference \cite{gordon2014probabilistic}, 2D image generation \cite{kulkarni2015picture}, and procedural modelling \cite{Ritchie2014QuicksandA}. The work with the closest domain to ours is Scenic \cite{fremont2018scenic}, a probabilistic language for generating top-down autonomous vehicle scenarios. ProbRobScene builds on the syntax of such languages, integrating ideas from convex geometry to support the inherently relational specifications in the domain of 3D robotic manipulation.

Another common approach to generating environments is using a probabilistic grammar. This technique has been used to create generative models of buildings \cite{chaudhuri2020learning}, traffic scenes \cite{devaranjan2020meta}, and even plants \cite{prusinkiewicz2012algorithmic}. However, while these methods excel at generating inputs with rich structure, phrasing a generator as a grammar makes it difficult to specify the inherently relational constraints \emph{between} objects that our domain requires.

Control improvisation \cite{fremont2017control} is another technique which uses formal methods to guide sampling from a generator. Here, one generates inputs by moving through an automaton. This technique works best in sequential domains where valid inputs can be expressed as visitation rules over graph traversals (for example, notes in a musical pattern). It is less well-suited to expressing spatial constraints seen in robotic environments.

A user can declaratively specify the task a robot must accomplish using a variety of different methods such as geometric constraints \cite{borghesan2015introducing}, temporal logic \cite{sadigh2016safe}, or expression graphs \cite{aertbelien2014etasl}. Such approaches complement our work here, as a user could specify both task and environment formally, then automatically verify their robotic system end to end (as in e.g., \cite{dreossi2019verifai, bohrer2018veriphy}). In particular, work on declaring task specifications using computer-aided design (CAD) semantics is promising \cite{somani2015constraint}, as such work mirrors ours by focusing on geometric relational constraints between objects. It might be possible to adapt such techniques such that the language for specifying both task and environment is shared.

If using ProbRobScene to generate training data for a learned robot model, our language can be seen as a form of data augmentation \cite{shorten2019survey}. Traditional methods of data augmentation like domain randomization \cite{tobin2017domain} rely on randomly perturbing an unstructured image space. More recent methods use generative adversarial networks (\textsc{gan}s) to generate data which looks similar to examples from a target domain \cite{zakharov2019deceptionnet}. Others try to directly learn user notions of qualitative \emph{functional relations} (e.g., support, protection, closeness) to generate environments which correspond to setups a user is likely to encounter in their chosen domain \cite{sjoo2011learning, kunze2014bootstrapping}. However, such methods lack a way for the user to directly inject explicit constraints into the augmentation process.

Many techniques focus on the problem of finding inputs for perception or control systems which will falsify a given task specification \cite{tuncali2018simulation,dreossi2019compositional, dreossi2018counterexample, annpureddy2011s}. In a sense, their focus is on solving a reverse problem: the input space is largely \emph{unconstrained}, but is so large that sampling it randomly is unlikely to yield examples which violate the task specification. To that end, techniques like covering arrays and error-tables are used to focus the search of the input space into areas likely to falsify the given specification. Such sampling techniques complement the work of ProbRobScene, and by combining the strengths of each, one could create a sampler capable of performing targeted searches of highly constrained input spaces

By having the user describe what range of environments their robot should operate in, ProbRobScene aligns with the goals of \emph{machine teaching} \cite{simard2017machine}. Under this paradigm, the focus is less on developing methods for improving the learning itself, but instead on providing tools for increasing the effectiveness of those teaching the system the correct behaviour. By using our environment specification to drive edge-case detection in simulation, our work also follows closely to the concept of \emph{envisioning} \cite{kunze2017envisioning}, where a robot with a high-level plan leverages simulations to predict whether its actions may have undesirable outcomes.

\section{Conclusion and Future Work}

In this paper, we merged ideas from probabilistic programming and convex geometry to create a specification language well-suited to the inherently relational environments found in robotic manipulation. We showed how the tight integration of these ideas yielded efficient sampling algorithms, and how such a language could be used for debugging a robotic controller in simulation.

In future work, we aim to focus on two main directions. Firstly, we aim to expand the range of relational specifiers available to the user to include concepts immediately related to robot mechanics (for example, using inverse kinematics to specify the region of space that is \emph{reachable} by a given robot arm). Secondly, we aim to develop ProbRobScene's usefulness as a tool for aiding machine learning in robotics. In particular, we intend to integrate techniques from counter-example guided learning \cite{dreossi2018counterexample} such that our language can be used to identify areas of the input space in which a given learned robot controller performs poorly, then provide targeted data augmentation to improve its performance.

\section*{ACKNOWLEDGMENT}
Thanks to the members of the Edinburgh Robust Autonomy and Decisions Group for their feedback and support. In particular, we are grateful for Advaith Sai's efforts to expand the range of examples available in the paper's code repository.

\bibliographystyle{unsrt}
\bibliography{references}

\end{document}